\documentclass[11pt]{article}
\usepackage[final]{acl}

\usepackage{times}
\usepackage{latexsym}

\usepackage[T1]{fontenc}

\usepackage[utf8]{inputenc}

\usepackage{microtype}

\usepackage{inconsolata}

\usepackage[normalem]{ulem} 
\usepackage{xcolor}   
\usepackage{ulem}     
\usepackage{tabularx}  
\usepackage{booktabs}  
\usepackage{hyperref}
\usepackage{algorithm}
\usepackage{algpseudocode}
\usepackage[utf8]{inputenc}
\usepackage{ulem}
\usepackage{color}
\usepackage{graphicx}
\usepackage{graphicx}
\usepackage{tabularx}      %
\usepackage{ragged2e}      
\usepackage{array}         %
\usepackage{arydshln} 
\usepackage{dashrule}
\usepackage{graphicx}%
\usepackage{multirow}%
\usepackage{amsmath,amssymb,amsfonts}%
\usepackage{amsthm}%
\usepackage{mathrsfs}%
\usepackage[title]{appendix}%
\usepackage{xcolor}%
\usepackage{textcomp}%
\usepackage{manyfoot}%
\usepackage{booktabs}%
\usepackage{algorithm}%
\usepackage{algorithmicx}%
\usepackage{algpseudocode}%
\usepackage{listings}%
\usepackage{acl}
\usepackage{marvosym}
\title{Named Entity Recognition in COVID-19 tweets with Entity Knowledge Augmentation}

\author{Xuankang Zhang \and Jiangming Liu\textsuperscript{\Letter} \\
        School of Information Science and Engineering, Yunnan University, China\\
        \textsuperscript{\Letter}corresponding author: jiangmingliu@ynu.edu.cn
        }

\begin{document}
\maketitle
\begin{abstract}
The COVID-19 pandemic causes severe social and economic disruption around the world, raising various subjects that are discussed over social media. Identifying pandemic-related named entities as expressed on social media is fundamental and important to understand the discussions about the pandemic. However, there is limited work on named entity recognition on this topic due to the following challenges: 1) COVID-19 texts in social media are informal and their annotations are rare and insufficient to train a robust recognition model, and 2) named entity recognition in COVID-19 requires extensive domain-specific knowledge. To address these issues, we propose a novel entity knowledge augmentation approach for COVID-19, which can also be applied in general biomedical named entity recognition in both informal text format and formal text format. Experiments carried out on the COVID-19 tweets dataset and PubMed dataset show that our proposed entity knowledge augmentation improves NER performance in both fully-supervised and few-shot settings. Our source code are publicly available:~\url{https://github.com/kkkenshi/LLM-EKA/tree/master}.
\end{abstract}

\section{Introduction}

The COVID-19 pandemic has led to significant social and economic upheaval globally, sparking various topics of conversation and debate on social media. Identifying pandemic-related entities mentioned on social media is crucial to highlight these discussions. However, there are limited named entity recognition data annotated with a focus on COVID-19 or public health research, making it difficult for pandemic-related analysis and NER model training~\cite{tjong-kim-sang-de-meulder-2003-introduction,pradhan-etal-2013-towards,strauss-etal-2016-results,hou-etal-2020-shot,jiang-etal-2022-annotating}.

\begin{figure*}[!tp]
    \centering
    \includegraphics[width=\textwidth]{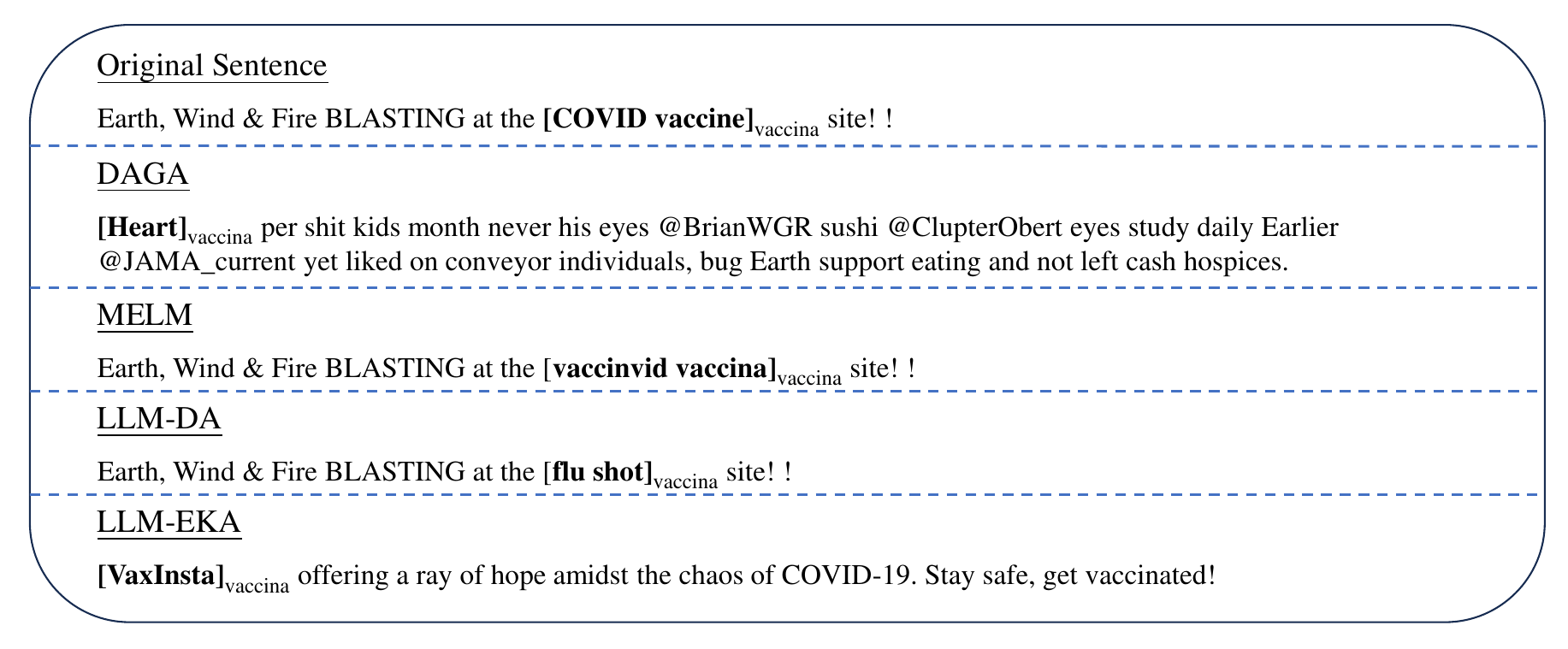}  
    \caption{Examples of various data augmentation methods in few-shot settings. The named entities are \textbf{[bold]}.}
    \label{fig:example} 
\end{figure*}

To address the challenges posed by the limited availability of annotated data, several data augmentation methods have been proposed to improve the performance of NER models. Traditional data augmentation methods, such as back-translation~\cite{sennrich2016improvingneuralmachinetranslation}, synonym replacement~\cite{wei-zou-2019-eda}, shuffle with segments~\cite{dai-adel-2020-analysis}, and the methods that rely on model-generated data~\cite{ding-etal-2020-daga,zhou-etal-2022-melm}, have been applied to enhance the diversity and robustness of training data. However, as shown in Figure~\ref{fig:example}, they either overlook the syntactic structure of the sentence or perform poorly in domain-specific contexts. For instance, DAGA~\cite{ding-etal-2020-daga} generates sentences with syntactic irregularities, which limits its effectiveness in NER tasks that depend on the precise syntactic and semantic structure of each token to accurately identify entities and their types. MELM~\cite{zhou-etal-2022-melm} tends to ignore the sentence context during substitution and exhibits limited scalability when adapting to new entities. LLM-DA~\cite{ye2024llm}, despite its flexibility, shows suboptimal performance in domain-specific scenarios, frequently generating vague or contextually inappropriate entity, especially when handling fine-grained entity types. 



Recent advancements in large language models (LLMs) have significantly improved performance in NER tasks across both general and domain-specific scenarios~\cite{huang2022copner,yang2022large,chen2023learning,meoni-etal-2023-large,sharma-etal-2023-paraphrase,bogdanov2024nuner}. These works inspire the development of more tailored augmentation frameworks that align LLMs with fine-grained entity knowledge. 
Building upon these advancements, the broader success of large language models in various NLP tasks has further expanded the possibilities for enhancing NER through more flexible and generalizable training paradigms. 
By leveraging their superior text representation capabilities, we propose a \textbf{LLM}-based \textbf{E}ntity \textbf{K}nowledge \textbf{A}ugmentation (\textbf{LLM-EKA}) to enrich the COVID-19-related knowledge of the models, which can also be applied into other domain-specific NER models. 

LLM-EKA consists of demonstration selection, entity augmentation, and instance augmentation, which effectively aligns LLMs to the domain-specific knowledge.
The demonstration selection aims to extract informative examples that are used as demonstrations for instance augmentation. The entity augmentation is applied to obtain domain-specific entities via large language models. The instance augmentation generates additional domain-specific training instances via prompts according to the selected demonstrations and augmented domain-specific entities.

The results of experiments carried out on both the METS-CoV benchmark and the BioRED benchmark show that the NER models equipped with the proposed LLM-EKA outperform the baseline models in the fully-supervised and few-shot settings. The main contributions of this work are summarized as follows:
\begin{itemize}
\item We investigate named entity recognition from medical research perspectives that contribute to public health concerns.
\item We propose a novel framework of entity knowledge augmentation for named entity recognition in COVID-19 tweets, which can also be applied into other domain-specific NER models.
\item Our final model, equipped with the proposed entity knowledge augmentation, achieves state-of-the-art results on benchmarks in both fully-supervised and few-shot settings. The codes will be released. 
\end{itemize}

\begin{figure*}[!tp]
    \centering
    \includegraphics[width=\textwidth]{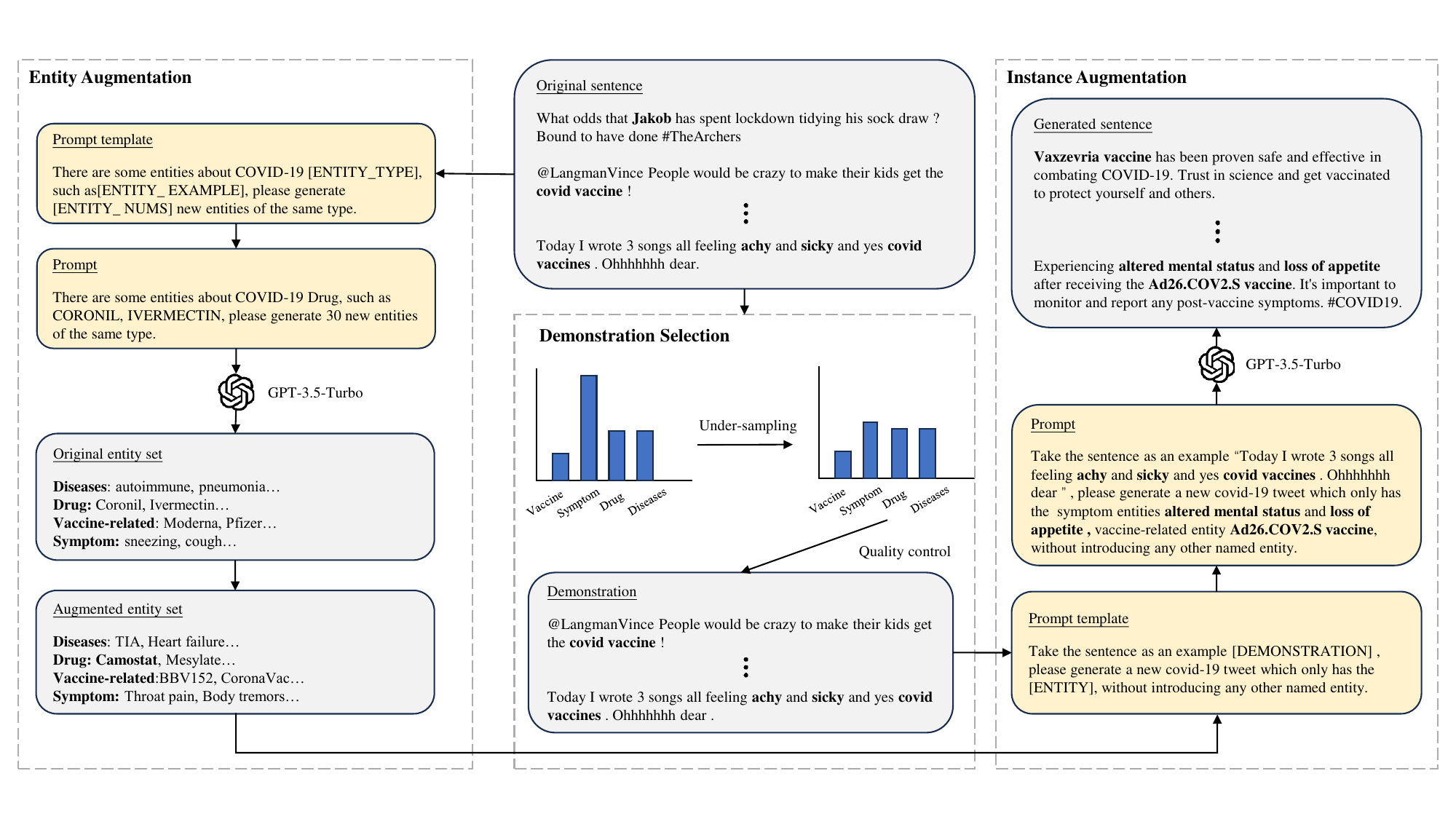} 
    \caption{Framework of LLM-based Entity Knowledge Augmentation.}
    \label{fig:framework}
\end{figure*}

\section{Related Work}
Named Entity Recognition is widely investigated as a conventional task in NLP. Recent work has focused on specific domains, which are often limited to small-scale training data.

\paragraph{Domain Transfer}
Domain transfer aims to alleviate the data scarcity issue by transferring knowledge from source domains to target domains. This line of research on NER enhances the generalization of models by aligning the entity knowledge of source domains with the target domains and enhances adaptability by mapping the entity label space between the source and target domains~\cite{daume-iii-2007-frustratingly}. Additionally, label embeddings are used as features to map label types across different domains, enabling cross-domain transfer~\cite{kim-etal-2015-new}. A label-aware dual transfer learning framework that uses a variant of the Maximum Mean Discrepancy (MMD) algorithm has been proposed~\cite{wang-etal-2018-label-aware}. Cross-Domain NER algorithms enhance model generalization by aligning entity knowledge from the source domain with the target domain~\cite{lee-etal-2018-transfer,lin-lu-2018-neural,yang-etal-2018-design}, and cross-domain language modeling and a novel parameter generation network have been utilized to perform knowledge transfer across domains and tasks~\cite{jia-etal-2019-cross}. A novel architecture for multi-domain NER has been presented that employs shared and private domain parameters along with multi-task learning to enhance model robustness across various text genres~\cite{wang-etal-2020-multi-domain-named}. Moreover, some related works have been conducted in few-shot settings~\cite{cui-etal-2021-template,ma-etal-2022-decomposed,ding-etal-2022-openprompt,chen-etal-2023-prompt,fang-etal-2023-manner}. However, these methods are based on the assumption that the label spaces between the source domain and target domain are aligned.

\paragraph {Data Augmentation} 
Data augmentation methods aim to directly expand the scale of training data to alleviate the data scarcity. Language generation models trained on labeled data have been used to generate samples for few-shot scenarios, achieving significant improvements in entity recognition using both supervised and unsupervised methods~\cite{ding-etal-2020-daga}. Similarly, simple token-level augmentation techniques have been proposed by adapting sentence-level task augmentation strategies, which have proven especially effective in enhancing NER performance in few-shot settings, particularly with small-scale training datasets~\cite{dai-adel-2020-analysis}. The label-tag misalignment issue in few-shot named entity recognition has been tackled by explicitly injecting NER labels into sentence context, and further improvements have been made in multilingual settings using code-mixing techniques~\cite{zhou-etal-2022-melm}. Moreover, LLMs have been leveraged to generate a large quantity of diverse and high-quality new data for NER~\cite{ye2024llm}. However, these methods suffer from a lack of domain-specific knowledge and are not well-suited to NER in domains that require extensive expertise.

\section{Methodology}
As shown in Figure \ref{fig:framework}, the framework of LLM-EKA consists of demonstration selection, entity augmentation, and instance augmentation. 

\subsection{Named Entity Recognition Model}
Named entity recognition is modeled as a sequence labeling task, where the input is a sequence of words, $[x_1, x_2, \ldots, x_n]$, and the output is a sequence of labels, $[y_1, y_2, \ldots, y_n]$, where $n$ is the length of the input sentence. We use the BIO labeling system for NER.

We use a pre-trained language model to obtain the hidden representation, which transforms the input tokens into their respective hidden representations, \( [h_1, h_2, \ldots, h_n] \). Subsequently, the hidden representation of each word is projected into a logit space,
$o_i = W h_i + b$,
where \( W \) and \( b \) are the learnable parameters. The logits \( o_i \) represent scores for each potential label associated with the word \( x_i \). The predicted label for each word \( y_i \) is then determined by applying the \( \arg\max \) function over the logits:
$y_i = \arg\max_{c} o_{i,c}$,
where \( c \) indexes the set of possible labels. The models are optimized by minimizing the cross-entropy loss:
\begin{equation}
\mathcal{L} = - \frac{1}{n} \sum_{i=1}^{n} \log p(y_i \mid x_i),
\end{equation}
where \( p(y_i \mid x_i) \) is the probability assigned to the true label by the model, which is obtained via a softmax operation over the logits.


\subsection{Demonstration Selection}

Given the training set, $S = \{ s_1, s_2, \dots, s_n \}$, where each instance $s_i$ has a set of entities $E_i$, we design different demonstration selection algorithms to fully-supervised settings and few-shot settings, respectively.

In fully-supervised settings, we perform under-sampling to filter out instances that contain the over-represented category, which balances the entity distribution, preventing model bias. A maximum threshold \( t \) on the number of entities per instance is taken to reduce complexity and mitigate noise, yielding a quality-controlled subset:

\begin{equation}
S_q = \left\{ s_i \in S \mid \lvert E_i \rvert \leq t \right\},
\end{equation}
where \( \lvert E_i \rvert \) denotes the number of entities in instance \( s_i \), where domain-specific entities have higher priority. The demonstration set for fully-supervised settings, $S_{\text{demo-full}}$ satisfies that each $s_i \in S_q $ should contain domain-specific entities:
\begin{align}
S_{\text{demo-full}} &= \left\{ s_i \in S_q \mid \right. \nonumber \\
&\left. \text{ $E_i$ has domain-specific entities} \right\}
\end{align}

In few-shot settings, we apply Algorithm~\ref{alg:few-shot} to sample a \(k\)-shot demonstration set. This algorithm ensures that each domain-specific entity appears at least $k$ times in the selected subset while preventing any entity type from being over-represented.\footnote{In our implementation, we set the tolerance parameter $\alpha$ to 1.3.}

\subsection{Entity Augmentation}
In few-shot NER tasks, particularly when dealing with domain-specific entities, entity augmentation plays a crucial role in addressing the scarcity of domain-specific annotated entities. 
We leverage large language models to expand the knowledge of domain-specific entities with prompts.
The prompt typically follows a template such as: ``There are some entities about COVID-19 [ENTITY TYPE] such as [ENTITY\_EXAMPLE]. Please generate [ENTITY\_NUMS] new entities of the same type.” where the [ENTITY\_TYPE] is the  domain-specific entity type, while [ENTITY\_EXAMPLE] is a concrete instance drawn from the training samples, [ENTITY\_NUMS] is the numerical quantity of new entities to be generated.

The approach encompasses two primary strategies for entity augmentation: straightforward strategy and iterative. In the straightforward approach, all available entity samples are input into the prompt at once, allowing the model to generate a set of new entities in a single operation. While this method is simple and efficient, it may lead to information overload if the prompt becomes overly lengthy, potentially impacting the quality of the generated entities due to diminished focus on individual entity characteristics.
To address this limitation, iterative strategy is introduced, where the entity samples are divided into smaller batches and fed into the model over multiple iterations. The strategy enable LLMs to concentrate on a narrow subset of entities at each step. 

\begin{algorithm}[!tp]
\small
\caption{Demonstration selection for \textit{k}-shot settings}
\label{alg:few-shot}
\begin{algorithmic}[1]
\renewcommand{\algorithmicrequire}{\textbf{Input:}}
\renewcommand{\algorithmicensure}{\textbf{Output:}}
\Require 
    Training set \( S = \{ s_1, \dots, s_n \} \)  

    Entity type set \( \mathcal{C} = \{ c_1, \dots, c_m \} \)  

    Tolerance \( \alpha \)
\Ensure 
    Demonstration set  $S_{\text{demo-few}}$  with \( k \leq C(c) \leq \alpha k,\ \forall c \in \mathcal{C} \)

\State Initialize \( S_{\text{demo-few}} \gets \emptyset \)
\State Initialize counter \( C(c) \gets 0,\ \forall c \in \mathcal{C} \)
\State Shuffle \( S \) randomly 

\For{each instance \( s_i \in S \)} 
        \State \( \Delta(c) \gets 0,\ \forall c \in \mathcal{C} \)

    \For{each entity type \( c \in s_i \)} 
        \If{ \( c \in \mathcal{C} \) }
            \State \( \Delta(c) \gets \Delta(c) + 1 \)
        \EndIf
    \EndFor

    \State Initialize \textbf{add} \(\gets \text{True} \)
    \For{each \( c \in \mathcal{C} \)}
        \If{ \( C(c) + \Delta(c) > \alpha k \) }
            \State \textbf{add} \(\gets \text{False} \), \textbf{break}
        \EndIf
    \EndFor

    \If{ \textbf{add} }
        \For{each \( c \in \mathcal{C} \)}
            \State \( C(c) \gets C(c) + \Delta(c) \)
        \EndFor
        \State \( S_{\text{demo-few}} \gets S_{\text{demo-few}} \cup \{ s_i \} \)
    \EndIf

    \If{ \( \forall c \in \mathcal{C},\ C(c) \geq k \) } 
        \State \textbf{break}
    \EndIf
\EndFor

\State \Return \( S_{\text{demo-few}} \)
\end{algorithmic}
\end{algorithm}

\begin{table*}[!htp]
    \centering
    \resizebox{\linewidth}{!}{
    \begin{tabular}{lcccccccc}
        \toprule
        \textbf{Model} & \textbf{Loc.} & \textbf{Org.} & \textbf{Per.} & \textbf{Sym.} & \textbf{Vacc.} & \textbf{Dise.} & \textbf{Drug} &\textbf{Avg} \\
        \midrule
        CRF & 76.37±0.62 & 54.64±2.08 & 64.43±1.59 & 74.05±0.56 & 84.85±0.82 & 73.61±0.44 & 77.34±1.60 & 71.58±0.54 \\
        BiLSTM + CRF & 79.25±0.59 & 62.39±0.97 & 72.41±0.44 & 79.14±0.51 & 88.72±0.62 & 74.89±1.28 & 79.60±0.71 & 76.38±0.22 \\
        BiLSTM + CLSTM + CRF&81.26±1.19 & 63.21±0.93 & 77.49±1.67 & 79.14±0.53 & 87.85±0.43 &75.61±0.76 &81.27±0.65 & 77.63±0.40 \\ 
        BiLSTM + CCNN + CRF &82.15±0.44 & 62.79±0.91 & 81.38±0.44 & 78.12±0.51 & 89.11±0.36 &76.12±0.76 &80.41±0.58 & 78.10±0.19 \\ 
        \midrule
        BART-large  &80.04±4.74 & 64.66±8.86 & 81.60±4.93 & 74.27±4.45 & 81.21±6.20 &71.24±1.90 &80.61±2.90&75.56±5.04\\
        BERT-large  &84.63±1.38 & 73.30±1.38 & 88.25±0.52 & 80.12±0.55 & 89.16±1.58 &76.52±1.11 &86.05±1.06&82.05±0.24\\
        RoBERTa-large  &85.85±2.12 & 73.78±0.72 & 86.79±0.44 & 81.32±0.67 & 90.42±1.12 &76.84±0.57 &86.79±0.78&82.55±0.27\\
        \midrule
        COVID-TWITTER-BERT & 85.68±0.92 & 76.27±0.64 & \textbf{\textcolor{red}{91.29±0.42}} & 81.85±0.53 & 90.44±0.94 & \uline{77.48±0.81} & 86.35±0.96& 83.88±0.20 \\
        ~~~~$+$DAGA &\uline{86.67±0.10} & 75.74±0.29 & 89.42±0.55 & 82.05±0.10 & 88.52±0.42 & 76.28±0.44 & 84.78±1.76 & 83.20±0.12  \\
        ~~~~$+$MELM &83.66±0.24 & 75.07±0.32 & 90.07±0.84 & 80.45±0.61 & 88.04±1.63 & 76.95±1.27 & 84.37±0.15 & 82.43±0.15  \\
        ~~~~$+$LLM-DA &85.78±0.66 & 75.16±0.51 & 89.82±0.95 & 80.88±0.38 & 88.90±0.70 & 76.75±0.72 & 86.49±0.55 & 82.96±0.17  \\
    \hdashline
        ~~~~$+$LLM-EKA-straight &86.33±0.57 & \textbf{\textcolor{red}{76.98±0.27}} & 90.62±0.58 & \uline{82.16±0.21} & \uline{90.46±1.05} & 77.46±0.88 & \uline{87.12±0.95} & \uline{84.08±0.10} \\
        ~~~~$+$LLM-EKA-iterative & \textbf{\textcolor{red}{86.78±1.00}} & \uline{76.69±0.53} & \uline{90.63±0.51} & \textbf{\textcolor{red}{82.35±0.43}} & \textbf{\textcolor{red}{92.05±0.52}} & \textbf{\textcolor{red}{77.72±0.80}} & \textbf{\textcolor{red}{87.14±0.45}} & \textbf{\textcolor{red}{84.32±0.10}}\\

        \bottomrule
    \end{tabular}
    }
    \caption{Results on test data in METS-CoV. LLM-EKA-straight represents the model enhanced with the straightforward strategy in entity augmentation. LLM-EKA-iterative represents the model enhanced with the iterative strategy in entity augmentations.
    The best scores are \textbf{\textcolor{red}{bold}} and the second best scores are \uline{underlined}.}
    
    \label{tab:evaluation-results}
\end{table*}

\subsection{Instance  Augmentation}
Building upon the expanded entity set obtained through Entity Augmentation, we introduce Instance Augmentation to generate diverse, contextually grounded textual instances that integrate the augmented entities while preserving contextual coherence and domain-specific relevance.  We generate new COVID-19 tweets by querying LLMs using a prompt template ``Take the sentence as an example [DEMONSTRATION], please generate a new COVID-19 tweet which only has the [ENTITY], without introducing any other named entity." The prompts have demonstration slot, [DEMONSTRATION], filled by sentences outputted from the demonstration selection, aiming to guide LLM to generate tweets with high quality and consistency in structure and style. The entity slot, [ENTITY], are filled by domain-specific entities obtained from the entity augmentation. 

Although template design and context constraints guide LLM to generate text that meets expectations, the inherent randomness of language models may still cause the generated results to deviate from the target. To alleviate the problem, we discard the sentences that contain entities outside the predefined entity set. This prevents the introduction of irrelevant or noisy entities. 

{
\begin{table*}[!htbp]
    \centering
    \renewcommand{\arraystretch}{1.2}
    \resizebox{\linewidth}{!}{
    \begin{tabular}{lccccccccc}
        \bottomrule
        \textbf{Model} & $k$ & \textbf{Loc.} & \textbf{Org.} & \textbf{Per.} & \textbf{Sym.} & \textbf{Vacc.} & \textbf{Dise.} & \textbf{Drug} & \textbf{Avg} \\
        \bottomrule
        COVID-TWITTER-BERT         & \multirow{6}{*}{5}  & 02.85±03.21 & 06.35±08.82 & 10.13±11.83 & 19.53±08.24 & 08.64±03.78 & 21.75±11.31 & 41.63±11.64 & 16.84±06.03 \\
        ~~~~$+$DAGA                &                         & 02.15±02.62 & 03.66±04.53 & 04.58±05.81 & 21.09±06.32 & 10.73±04.85 & 23.35±11.20 & 39.64±15.18 & 15.50±04.48 \\
        ~~~~$+$MELM                &                         & 13.46±08.37 & 16.66±09.64 & 19.78±17.51 & 27.05±04.58 & 13.20±04.95 & 30.66±05.74 & 52.05±03.64 & 24.54±04.31 \\
        ~~~~$+$LLM-DA              &                         & 23.21±15.32 & 21.03±04.90 & 10.76±06.47 & \uline{44.23±11.00} & 27.13±08.37 & 38.84±02.01 & 59.93±03.78 & 34.55±04.42 \\
        \hdashline
        ~~~~$+$LLM-EKA             &                         & \uline{47.67±02.82} & \uline{32.60±04.65} & \textbf{\textcolor{red}{41.88±11.08}} & 36.24±04.60 & \textbf{\textcolor{red}{37.94±01.11}} & \uline{44.65±03.40} & \uline{64.52±05.25} & \uline{41.73±02.61} \\
        ~~~~~~~~w/self-verification    &                         & \textbf{\textcolor{red}{48.18±07.40}} & \textbf{\textcolor{red}{32.63±08.76}} & \uline{36.60±14.78} & \textbf{\textcolor{red}{46.23±5.74}} & \uline{35.55±05.59} & \textbf{\textcolor{red}{44.69±05.52}} & \textbf{\textcolor{red}{65.78±03.47}} & \textbf{\textcolor{red}{43.65±06.91}} \\
        \hline
        
        COVID-TWITTER-BERT                & \multirow{6}{*}{10} & 21.09±15.10 & 09.08±07.17 & 24.56±19.22 & 28.13±09.32 & 25.38±14.38 & 34.94±21.55 & 51.23±19.80 & 27.19±14.24 \\
        ~~~~$+$DAGA                &                         & 31.00±09.79 & 12.52±07.68 & 33.58±09.71 & 39.73±05.02 & 33.01±08.85 & 52.02±01.98 & 62.65±02.98 & 37.82±01.28 \\
        ~~~~$+$MELM                &                         & 43.48±04.67 & 28.27±00.70 & 53.23±07.26 & 34.89±02.22 & 28.59±00.96 & 50.67±00.02 & 61.37±00.55 & 41.21±01.29 \\
        ~~~~$+$LLM-DA              &                         & 54.08±04.38 & 33.32±02.75 & 48.29±10.68 & \textbf{\textcolor{red}{45.86±03.22}} & 38.85±06.02 & 49.40±04.67 & 65.90±02.76 & 46.68±02.09 \\
        \hdashline
        ~~~~$+$LLM-EKA             &                         & 42.69±01.07 & 35.94±00.41 & \textbf{\textcolor{red}{59.72±02.47}} & 44.21±00.99 & \textbf{\textcolor{red}{58.00±01.74}} & 51.70±00.34 & 63.65±00.09 & 49.54±00.57 \\
        ~~~~~~~~w/ self-verification    &                         & \textbf{\textcolor{red}{59.96±03.58}} & \textbf{\textcolor{red}{38.49±02.17}} & 50.33±07.91 & 44.77±02.35 & 57.19±04.79 & \textbf{\textcolor{red}{53.08±02.93}} & \textbf{\textcolor{red}{67.13±02.85}} & \textbf{\textcolor{red}{50.50±02.32}} \\

        \hline
        
        COVID-TWITTER-BERT                & \multirow{6}{*}{20} & 45.48±08.50 & 36.02±02.15 & 48.06±04.16 & 50.31±02.47 & 50.37±05.30 & \uline{59.16±01.60} & \uline{68.27±01.19} & 50.13±02.70 \\
        ~~~~$+$DAGA                &                         & 53.59±04.19 & 33.98±03.98 & 55.73±03.56 & 53.30±02.89 & 46.36±03.99 & 52.87±02.15 & 66.18±03.76 & 51.66±02.06 \\
        ~~~~$+$MELM                &                         & 61.19±04.52 & \uline{45.33±02.90} & \textbf{\textcolor{red}{76.47±05.76}} & \textbf{\textcolor{red}{55.16±02.07}} & 50.99±02.15 & 54.45±01.41 & 66.60±03.52 & \uline{56.61±00.81} \\
        ~~~~$+$LLM-DA              &                         & \uline{65.72±02.47} & 45.21±02.87 & 58.36±03.18 & 50.56±00.30 & \uline{56.98±04.06} & 54.24±06.00 & 66.79±00.17 & 55.28±02.59 \\
        \hdashline
        ~~~~$+$LLM-EKA             &                         & 57.41±01.00 & 42.84±01.49 & \uline{66.96±02.24} & \uline{54.75±02.11} & 55.47±04.79 & \textbf{\textcolor{red}{59.25±00.90}} & 65.83±00.29 & 56.60±00.33 \\
        ~~~~~~~~w/ self-verification   &                         & \textbf{\textcolor{red}{68.75±01.17}} & \textbf{\textcolor{red}{45.74±02.12}} & 63.65±03.24 & 53.23±03.11 & \textbf{\textcolor{red}{63.04±03.84}} & 57.65±02.38 & \textbf{\textcolor{red}{70.66±02.27}} & \textbf{\textcolor{red}{58.72±01.18}} \\
        \toprule

    \end{tabular}
    }
    \caption{Results of COVID-TWITTER-BERT with various data augmentation methods on METS-CoV test data in $k$-shot setting. The best scores are \textbf{\textcolor{red}{bold}} and the second best scores are \uline{underlined}.
    }
     
    \label{tab:METS-fewshot-results}
\end{table*}
}
\section{Experiments}
We conduct experiments in fully-supervised settings and few-shot settings by comparing our methods with recent data augmentations. Additionally, we show the performance of recent LLM-based methods on COVID-19 and biomedical NER tasks to demonstrate the effect of our proposed methods.

\subsection{Data}
The experiments are conducted on the METS-CoV benchmark which contains 7000 tweets with 7 entity types~\cite{zhou2022mets}, including 4 medical entity types: disease, drug, symptom and vaccine, and 3 general entity types: person, location and organization. In order to demonstrate that our model can be generalized to other biomedical NER task, we also conduct the experiments on BioRED benchmark~\cite{Luo_2022}, a set of 600 PubMed abstracts including 6 entity types: disease or phenotypic feature, chemical entity, gene or gene product, organism taxon, sequence variant, cell line.


\subsection{Baselines}
We compare our models to traditional NER methods and the models based on pre-trained language model. The traditional NER models include CRF, BiLSTM + CRF, BiLSTM + CLSTM + CRF, and BiLSTM + CCNN + CRF, where CLSTM and CCNN are the modules of encoding character information with LSTM and CNN, respectively. For pre-trained language models, we utilize BERT-large, RoBERTa-large, and BART-large. Additionally, in order to make the models with ability of handling tweets, we additionally take several recent data augmentation methods to base NER models. In addition, we include comparisons with LLM-based NER methods, which leverage the generative capabilities of large language models for few-shot name entity recognition.
The data augmentation are summarized as follows:
\begin{itemize}
    \item DAGA~\cite{ding-etal-2020-daga} linearize each labeled sentence and training a lightweight single‑layer LSTM to capture their joint distribution via next‑token prediction, which is used to generate high‑quality synthetic annotated data.\footnote{\url{https://ntunlpsg.github.io/project/daga/}}
    \item MELM~\cite{zhou-etal-2022-melm} predicts masked entity tokens based on their corresponding NER tags to generate diverse and label-consistent synthetic sentences.\footnote{\url{https://github.com/RandyZhouRan/MELM/}}
    \item LLM-DA~\cite{ye2024llm}  leverages large language models through structured prompting and multi-level augmentation by contextual rewriting, entity replacement, and noise injection.
\end{itemize}
COVID-TWITTER-BERT~\cite{muller2023covid} and PubMedBERT~\cite{gu2021domain} are adopted as base NER models for METS-CoV and BioRED, respectively.

\subsection{Settings}
We apply GPT-3.5-turbo for the knowledge argumentation with the temperature of 1, fully leveraging the diversity generation capabilities. We set the batch size to 8 and employ the AdamW optimizer with a learning rate of 3e-5. The models are trained in 100 epochs. Micro F1 scores are used as our evaluation metrics. We set up 5-shot, 10-shot, and 20-shot settings for few-shot experiments. All the models are trained on one 24G GPU. 



{
\begin{table*}[!tp]
    \centering
    \renewcommand{\arraystretch}{1.2}
    \resizebox{\linewidth}{!}{
    \begin{tabular}{lcccccccc}
        \bottomrule
        \textbf{Model} & $k$ & \textbf{Cell.} & \textbf{Chem.} & \textbf{Dise.} & \textbf{Gene.} & \textbf{Orga.} & \textbf{Sequ.} & \textbf{Average} \\
        \midrule
        PubMedBERT      & \multirow{6}{*}{5}  & 58.24±09.42 & 42.05±23.65 & 07.80±06.10 & 44.67±15.91 & 30.19±11.24 & 40.44±10.43 & 35.86±08.48 \\
        ~~~~$+$DAGA      &                         & 06.62±06.58 & 32.35±06.66 & 10.22±07.88 & 25.53±05.31 & 35.05±11.49 & 40.93±08.77 & 26.05±05.42 \\
        ~~~~$+$MELM      &                         & 32.19±12.11 & 47.25±12.45 & 21.84±03.85 & 55.10±05.43 & 49.78±07.89 & 49.13±08.62 & 43.90±03.64 \\
        ~~~~$+$LLM-DA    &                         & 52.43±09.36 & 68.49±04.13 & 27.75±08.44 & 66.39±02.23 & 19.03±15.09 & 44.30±06.41 & 51.53±02.24 \\
        \hdashline
        ~~~~$+$LLM-EKA   &                         & \uline{66.14±07.07} & \uline{70.42±02.97} & \uline{42.86±02.16} & \uline{71.97±00.41} & \uline{66.71±09.92} & \textbf{\textcolor{red}{56.65±05.14}} & \uline{62.79±00.61} \\
        ~~~~~~~~w/ self-verification &                 & \textbf{\textcolor{red}{67.14±06.72}} & \textbf{\textcolor{red}{71.27±02.48}} & \textbf{\textcolor{red}{47.13±03.44}} & \textbf{\textcolor{red}{72.86±00.87}} & \textbf{\textcolor{red}{69.07±06.91}} & \uline{54.05±01.66} & \textbf{\textcolor{red}{64.33±01.15}} \\
        \hline\rule{0pt}{8pt}

        PubMedBERT      & \multirow{6}{*}{10} &  \textbf{\textcolor{red}{68.77±03.13}} & 67.39±04.39 & 23.30±04.01 & 57.96±10.22 & 53.48±07.22 & 47.18±11.93 & 51.83±04.22 \\
        ~~~~$+$DAGA      &                         & 47.22±07.29 & 62.19±01.59 & 22.40±04.41 & 52.37±04.02 & 53.20±10.52 & 39.85±02.89 & 46.85±03.28 \\
        ~~~~$+$MELM      &                         & 48.89±08.95 & 71.75±01.55 & 30.37±02.46 & 63.23±03.18 & 54.23±08.01 & 48.67±05.69 & 53.66±02.23 \\
        ~~~~$+$LLM-DA    &                         & 61.20±09.67 & \uline{74.64±01.51} & 37.61±01.86 & 72.00±05.80 & 61.84±06.60 & 47.97±10.39 & 61.22±02.44 \\
        \hdashline
        ~~~~$+$LLM-EKA   &                         & \uline{68.19±12.54} & 74.04±01.82 & \uline{54.24±02.68} & \uline{73.45±02.57} & \textbf{\textcolor{red}{77.54±04.56}} & \uline{51.92±03.13} & \uline{67.51±00.80} \\         
        ~~~~~~~~w/ self-verification &                 & 66.96±02.97 & \textbf{\textcolor{red}{76.14±01.57}} & \textbf{\textcolor{red}{57.23±01.50}} & \textbf{\textcolor{red}{74.73±01.57}} &\uline{76.68±02.68} & \textbf{\textcolor{red}{54.11±00.74}} & \textbf{\textcolor{red}{69.05±00.74}} \\

        \hline
        
        PubMedBERT      & \multirow{6}{*}{20} & 54.74±03.32 & 75.95±00.18 & 47.98±00.16 & 70.32±00.85 & \textbf{\textcolor{red}{85.76±01.18}} & \uline{59.14±01.46} & 66.27±00.40 \\
        ~~~~$+$DAGA      &                         & 52.21±14.48 & 75.14±02.12 & 41.32±01.80 & 67.95±01.83 & \uline{83.29±01.95} & 50.92±01.75 & 62.84±00.47 \\
        ~~~~$+$MELM      &                         & 58.87±18.83 & 77.95±01.49 & 52.20±02.23 & 72.32±04.72 & 80.53±01.03 & 55.79±03.06 & 67.73±02.75 \\
        ~~~~$+$LLM-DA    &                         & \uline{70.00±12.20} & \textbf{\textcolor{red}{78.24±01.44}} & 45.73±01.47 & 74.68±03.13 & 82.34±01.54 & 55.93±01.03 & 67.47±00.96 \\
        \hdashline
        ~~~~$+$LLM-EKA   &                         & \textbf{\textcolor{red}{73.93±01.33}} & 71.93±01.77 & \textbf{\textcolor{red}{58.56±01.38}} & \uline{78.51±01.31} & 79.64±04.06 & \textbf{\textcolor{red}{62.49±01.82}} & \uline{70.79±00.18} \\
        ~~~~~~~~w/ self-verification &                 & 67.81±02.67 & \uline{77.98±00.62} & \uline{55.32±00.79} & \textbf{\textcolor{red}{82.21±01.07}} & 75.09±05.35 & \textbf{\textcolor{red}{62.49±04.66}} & \textbf{\textcolor{red}{72.24±00.04}} \\
        \toprule
    \end{tabular}
    }
    \caption{Results of PubMedBERT with various data augmetations on BioRED test data in $k$-shot setting. The best scores are \textbf{\textcolor{red}{bold}} and the second best scores are \uline{underlined}.}
    \label{tab:BIO-fewshot-results}
\end{table*}
}

\subsection{Fully-supervised}

Table \ref{tab:evaluation-results} shows the results on the test data across different models. Pre-trained language models BERT and RoBERTa significantly outperforms the traditional LSTM models with CRF. COVID-TWITTER-BERT outperforms RoBERTa because it has the ability to represent the COVID-19 tweets with help of the pre-training on tweets. Equipped with our proposed entity knowledge augmentation, the final model achieve the best results.
\paragraph{Incremental Augmentation}
We experiment with two augmentation approaches: LLM-EKA-straight and LLM-EKA-iterative. As shown in Table~\ref{tab:evaluation-results}, both LLM-EKA approaches outperforms baselines, achieving an average score of 84.08 and 84.32, respectively. 
By incrementally refining entity knowledge, LLM-EKA-iterative reduces noise and prevents LLMs information overload, leading to enhanced performance and better handling of long-distance dependencies.

\paragraph{Entity Type}
We perform a fine-grained analysis of the performance of different entity types across models. LLM-EKA-iterative show significant improvements in recognizing domain-specific entities, particularly for drugs (87.14 F1) and vaccine-related entities (92.05 F1). This suggests that our knowledge augmentation methods, especially with iterative refinement, enhance the model's ability to discern and accurately classify entities within the context of COVID-19 tweets. 


\subsection{Few-shot}

Table~\ref{tab:METS-fewshot-results} and Table \ref{tab:BIO-fewshot-results} 
show the performances of various methods in k-shot settings, respectively.\footnote{We take LLM-EKA-iterative as our model LLM-EKA in few-shot experiments.} In few-shot settings, our method consistently surpasses existing approaches across multiple datasets. On the METS-CoV dataset, which includes both general and domain-specific entities, our approach improves performance by 10–15 points in the 5-shot and 10-shot settings, demonstrating its effectiveness in capturing domain-specific knowledge. Similarly, on the BioRED dataset, which covers diverse biomedical entity types, our method achieves notable gains, outperforming prior methods by a significant margin, particularly in extremely low-resource conditions.

\paragraph{Data Domains}
METS-CoV contains informal, user-generated text with abbreviations and conversational language, while BioRED consists of well-structured biomedical literature with formal terminology. The difference impacts performance gains. in BioRED, our method achieves substantial improvements. In contrast, METS-CoV informal style introduces greater linguistic variability, making augmentation less effective. Moreover, METS-CoV includes three general entity types (person, organization, location), which are more common across datasets and less domain-specific, further affecting augmentation consistency.

\paragraph{Self-verification}
Self-verification mechanism plays a crucial role in improving model robustness in few-shot settings by filtering out domain-irrelevant augmentations (e.g., non-COVID-19-related disease entities such as \textit{Brucellosis} or drug entities like \textit{Chloroquine}) while enforcing precision in domain-specific terminology. 
However, as training data scales to 20-shot, the increased lexical diversity (e.g., formal names like \textit{BNT162b2} and regional variants) enables the mechanism to better distinguish noise from valid expressions.

\begin{table}[!tp]
\centering
\resizebox{\linewidth}{!}{
\begin{tabular}{lcccc}
\toprule
\multirow{2}{*}{\textbf{Model}} & \multicolumn{2}{c}{\textbf{5-shot}} & \multicolumn{2}{c}{\textbf{10-shot}} \\
     & \textbf{METS-CoV} & \textbf{BioRED} & \textbf{METS-CoV} & \textbf{BioRED} \\
\midrule

GPT-NER & 23.96  & 30.81 & \uline{35.88} & 35.68 \\
RT   & \uline{34.59}  & \uline{32.43} & 35.54 & \uline{39.64}\\
\hdashline
ours & \textbf{\textcolor{red}{43.65}} & \textbf{\textcolor{red}{64.33}} & \textbf{\textcolor{red}{50.50}} & \textbf{\textcolor{red}{69.05}}\\
\bottomrule
\end{tabular}
}
\caption{Results on METS-CoV and BioRED test data in 5/10-shot settings. The best scores are \textbf{\textcolor{red}{bold}} and the second best scores are \uline{underlined}.}
\label{tab:llm_microf1_fewshot}
\end{table}

\begin{table*}[!tp]
\centering
\begin{tabularx}{\textwidth}{l>{\justifying\noindent\arraybackslash}X}
\toprule
\textbf{Method} & \textbf{Result} \\
\midrule
Case A & 
Having intense \textbf{[allergies]}{\textsubscript{symptom}} and waking up every morning with a \textbf{[sore throat]}{\textsubscript{symptom}} and hoping by noon it goes away bc it was just from drainage and not COVID. \\
\midrule
LLM-DA & 
Having intense allergies and waking up every morning with a \textbf{[sore throat]}{\textsubscript{symptom}}  and hoping by noon it goes away bc it was just from drainage and not COVID. \\
LLM-EKA & 
Having intense \textbf{[allergies]}{\textsubscript{symptom}} and waking up every morning with a \textbf{[sore throat]}{\textsubscript{symptom}} and hoping by noon it goes away bc it was just from drainage and not COVID. \\
\midrule
\midrule
Case B & 
Um, actually, we do vaccinate healthy people against diseases— \textbf{[Polio]}{\textsubscript{disease}}, \textbf{[Tetanus]}{\textsubscript{disease}}, \textbf{[MMR]}{\textsubscript{disease}}, \textbf{[pneumonia]}{\textsubscript{disease}}, to name a few—everyone has the potential to be at risk from COVID.\\
\midrule
LLM-DA & Um, actually, we do vaccinate healthy people against diseases—Polio, \textbf{[Tetanus]}{\textsubscript{drug}}, MMR, \textbf{[pneumonia]}{\textsubscript{disease}}, to name a few—everyone has the potential to be at risk from COVID.\\
LLM-EKA & 
Um, actually, we do vaccinate healthy people against diseases— \textbf{[Polio]}{\textsubscript{disease}}, \textbf{[Tetanus]}{\textsubscript{disease}}, \textbf{[MMR]}{\textsubscript{disease}}, \textbf{[pneumonia]}{\textsubscript{disease}}, to name a few—everyone has the potential to be at risk from COVID.\\
\bottomrule
\end{tabularx}
\caption{Examples of NER in 5-shot setting.}
\label{tab:case-study}
\end{table*}

\subsection{Comparison to LLM-based Models}
The LLM-based models are summarized as follows:
\begin{itemize}
    \item GPT-NER~\cite{wang2023gpt} reformulates NER as generation, prompt an LLM to wrap entities with special tokens, retrieve entity-level or sentence-level kNN demonstrations for few-shot guidance, and apply self-verification to reduce hallucination.\footnote{\url{https://github.com/ShuheWang1998/GPT-NER}}
    \item RT~\cite{Li2023HowFI} enhances few-shot medical NER by retrieving relevant demonstrations via entity-aware KNN, followed by step-by-step entity prediction using chain-of-thought prompts.\footnote{\url{https://github.com/ToneLi/RT-Retrieving-and-Thinking}}
\end{itemize}

Table~\ref{tab:llm_microf1_fewshot} shows the results of LLM-based NER models in few-shot settings.
Although recent LLM-based NER methods such as GPT-NER and RT demonstrate promising potential in few-shot scenarios, their practical effectiveness on challenging benchmarks like METS-CoV and BioRED reveals several critical limitations. 
In 5-shot setting,
GPT-NER achieves micro F1 scores of 23.96 on METS-CoV and 30.81 on BioRED, while RT slightly outperforms it with 34.59 and 32.43, respectively. However, both models lag significantly behind our methods, which achieves 43.65 on METS-CoV and 64.33 on BioRED in the same setting. In 10-shot setting, the performance gap becomes more significant. It shows that general LLM hardly captures the domain-specific knowledge, which is the focus of our proposed methods.

\subsection{Case Study}
Table~\ref{tab:case-study} shows the outputs across LLM-DA and LLM-EKA on COVID-19 tweets dataset.  LLM‑DA fails to recognize certain entity and assigns incorrect types. In Case A, it fails to detect the standalone symptom entity \textit{allergies}, while by incorporating entities knowledge-enhanced prompts, our method achieves superior recognition of domain-specific terminology. 
In Case B, LLM‑DA not only fails to label \textit{Polio} and \textit{MMR} but also misclassifies \textit{Tetanus} as a drug, while LLM‑EKA effectively leverages entities contextual semantics and structural information to pinpoint the boundary and type of entities with high precision.

\section{Conclusion}
We present a novel LLM-based entity knowledge augmentation for named entity recognition in COVID-19 tweets for public health research. 
LLM-EKA leverages the sophisticated contextual reasoning capabilities and extensive knowledge base of LLMs to augment entity knowledge and improve recognition performance. Our proposed methods are model-agnostic and can be adapted to enhance NER models in other biomedical subdomains.  The experimental results show that our proposed data augmentation methods outperform previous work in fully-supervised and few-shot settings. In few-shot settings, our LLM-EKA is capable of addressing the challenges associated with scarce annotated data and the need for domain-specific expertise.

\section*{Limitations}
LLM-EKA current applicability is limited to the biomedical domain and the framework relies on a fixed base model, meaning its performance is inherently constrained by the capabilities and representation limits of the underlying encoder. Furthermore, our use of proprietary LLMs via API introduces dependence on third-party services.
\section*{Acknowledgements}
The work is supported by National Natural Science Foundation of China (grant No. 62566064) and Yunnan Fundamental Research Projects (grant No.\ 202401CF070189).
\bibliography{acl_latex}




\end{document}